\documentclass[letterpaper]{article} 
\usepackage{aaai2026}  
\usepackage{times}  
\usepackage{helvet}  
\usepackage{courier}  
\usepackage[hyphens]{url}  
\usepackage{graphicx} 
\usepackage{natbib}  
\usepackage{caption} 
\usepackage{algorithm}
\usepackage{algorithmic}
\usepackage{multirow}
\usepackage{makecell}
\usepackage{tipa}
\usepackage{utfsym}
\usepackage{subcaption}
\usepackage{graphicx}
\usepackage{amsmath}
\usepackage{csquotes}

\usepackage{newfloat}
\usepackage{listings}
\DeclareCaptionStyle{ruled}{labelfont=normalfont,labelsep=colon,strut=off} 
\floatstyle{ruled}
\newfloat{listing}{tb}{lst}{}
\floatname{listing}{Listing}
\title{Text2Lip: Progressive Lip-Synced Talking Face Generation from Text via Viseme-Guided Rendering}
\author{
  Xu Wang, Shengeng Tang\thanks{Corresponding author.}, Fei Wang, Lechao Cheng, Dan Guo, Feng Xue, Richang Hong
}
\affiliations{
    Hefei University of Technology\\

    wangxu2002@mail.hfut.edu.cn, tangsg@hfut.edu.cn
}

\usepackage{bibentry}

\begin{document}
\nocopyright

\maketitle

\begin{abstract}
Generating semantically coherent and visually accurate talking faces requires bridging the gap between linguistic meaning and facial articulation. Although audio-driven methods remain prevalent, their reliance on high-quality paired audio visual data and the inherent ambiguity in mapping acoustics to lip motion pose significant challenges in terms of scalability and robustness.
To address these issues, we propose \textbf{Text2Lip}, a viseme-centric framework that constructs an interpretable phonetic-visual bridge by embedding textual input into structured viseme sequences. These mid-level units serve as a linguistically grounded prior for lip motion prediction. Furthermore, we design a progressive viseme-audio replacement strategy based on curriculum learning, enabling the model to gradually transition from real audio to pseudo-audio reconstructed from enhanced viseme features via cross-modal attention. This allows for robust generation in both audio-present and audio-free scenarios. Finally, a landmark-guided renderer synthesizes photorealistic facial videos with accurate lip synchronization. Extensive evaluations show that Text2Lip outperforms existing approaches in semantic fidelity, visual realism, and modality robustness, establishing a new paradigm for controllable and flexible talking face generation. Our project homepage is https://plyon1.github.io/Text2Lip/.

\end{abstract}

\section{Introduction}
\begin{figure}[t]
  \centering
  \includegraphics[width=\linewidth]{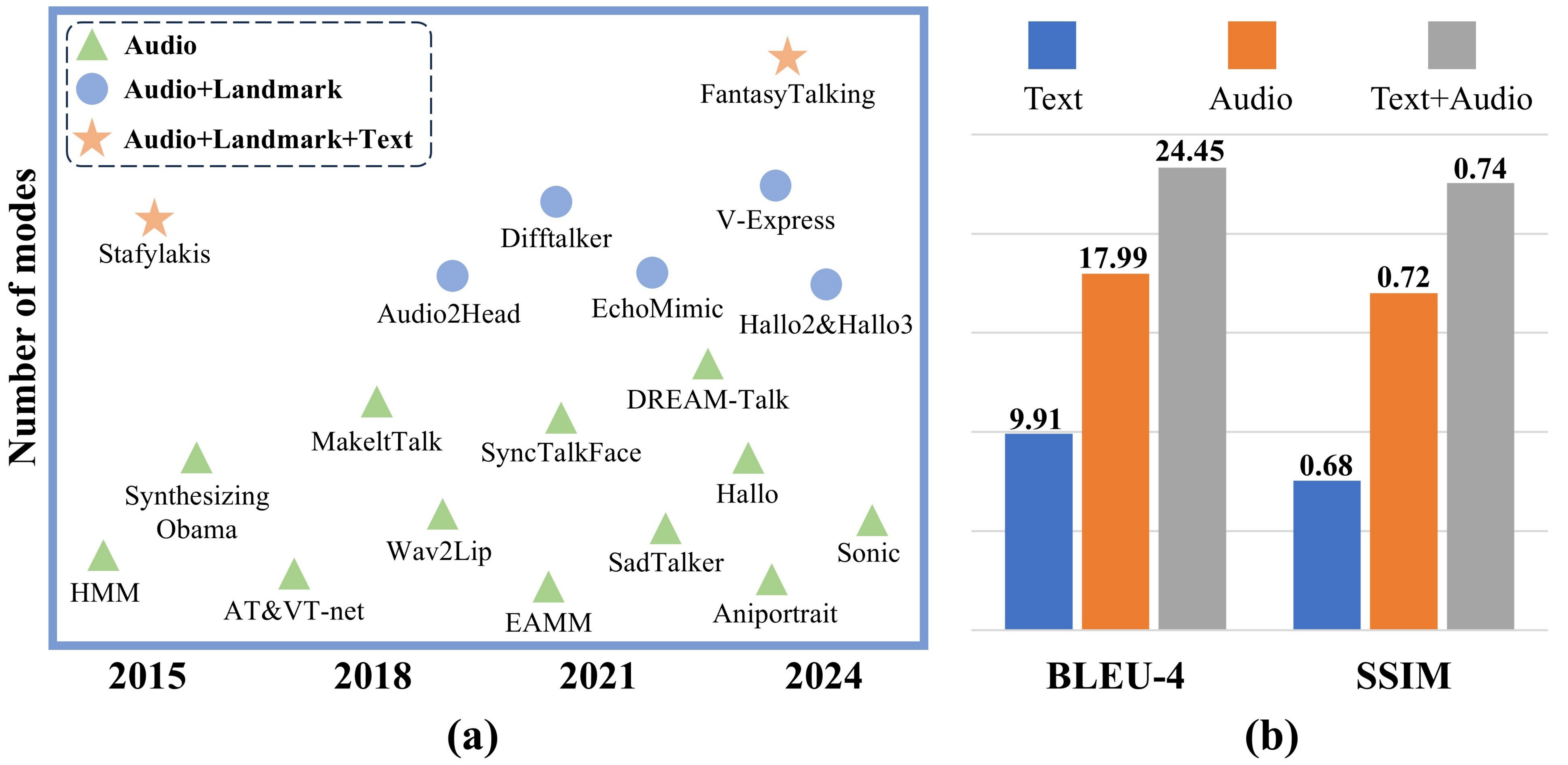}
  \caption{
  Modality usage in talking face generation. (a) Existing methods predominantly rely on audio inputs to drive lip motion. However, aligned audio-visual data is often scarce or costly to obtain. (b) Quantitative comparisons show that audio outperforms text-only input in fidelity, while combining both modalities yields further improvements. These observations motivate our viseme-centric approach, which leverages the accessibility and structure of text to bridge linguistic semantics and facial motion—enabling robust synthesis even in the absence of audio.
  }
  \label{overall}
\end{figure}

Talking face generation aims to synthesize photorealistic facial videos synchronized with speech or textual content, enabling applications in virtual avatars~\cite{tang2025discrete, wang2025signaligner}, accessible assistive communication~\cite{tangslt2022, Guo_Tang}, and human-computer interaction~\cite{tang2024GCDM, tang2025sign}. Most existing approaches adopt an \textit{audio-driven} paradigm, where acoustic features directly guide lip motion via 3D deformable models~\cite{xu2024vasa, wei2024aniportrait} or facial landmarks~\cite{zhou2020makelttalk, wang2024v}. While audio provides rich temporal and prosodic cues for driving lip motion, obtaining high-quality paired audio-video data remains costly and error-prone. In contrast, text as input is more flexible and accessible, making it an attractive alternative, especially in low-resource or privacy-sensitive scenarios. Apart from this, directly relying on audio overlooks the intrinsic coupling among \textit{text}, \textit{phonetics}, and \textit{visual articulation} that underlies human communication. This leads to a critical limitation: speech-driven models often learn ambiguous mappings from audio to lip shapes. For example, different semantic expressions such as “bad boy” and “bat boat” may correspond to nearly identical lip motions. This ambiguity compromises semantic precision and visual expressiveness, particularly when emotional clarity or linguistic fidelity is essential. As illustrated in Figure~\ref{overall}, these challenges motivate us to rethink modality design. We explore how the inherent structure of language, when grounded through phonetic and visemic representations, can effectively guide facial animation, even in the absence of audio.


To address this, we introduce \textbf{Text2Lip}, a viseme-aware framework that explicitly models the linguistic-phonetic-visual hierarchy. Instead of relying solely on audio, we convert input text into interpretable viseme sequences via word-to-phoneme-to-viseme mapping, which serve as semantically grounded priors for facial motion synthesis. This design disambiguates visually similar phonemes and enhances semantic-lip alignment.

Built upon this foundation, we develop a curriculum-based \textit{Progressive Viseme-Audio Replacement} strategy. During training, we gradually transition from real audio to text-derived viseme embeddings, enabling our model to reconstruct pseudo-audio features through cross-modal attention. This approach not only allows seamless adaptation to audio-free settings but also enhances robustness against noisy or missing speech. To ensure high-quality visual synthesis, we adopt a landmark-guided \textit{Photorealistic Video Rendering} module adapted from EchoMimic~\cite{chen2025echomimic}, which converts the predicted landmarks and optional (real or pseudo) audio into temporally smooth, lip-synced talking face videos. Our contributions are summarized as follows:
\begin{itemize}
\item We propose a novel viseme-centric framework that systematically aligns linguistic, phonetic, and visual representations, enabling more interpretable and controllable talking face generation.
\item We introduce a curriculum-based viseme-audio replacement strategy that facilitates flexible modality handling, supporting both audio-driven and audio-free generation scenarios.
\item Our approach achieves state-of-the-art performance in lip synchronization, semantic consistency, and cross-modal robustness, demonstrated on multiple benchmarks under varied conditions.
\end{itemize}

\section{Related work}

\subsection{Audio-Driven Talking Face Generation}
Audio-driven talking face generation is a long-standing core focus in audiovisual synthesis. Traditional methods employ explicit 3D representations, such as 3DMM-based approaches~\cite{zhang2023sadtalker, zhang2023dream, wei2024aniportrait, xu2024vasa}, which offer controllability but struggle to capture fine-grained facial dynamics. End-to-end models~\cite{chen2025echomimic, xu2024hallo, ferdowsifard2021loopy} bypass 3D intermediates by directly mapping audio to pixels, improving realism and synchronization. Extensions like Hallo2/3~\cite{cuihallo2, cui2025hallo3} enrich expression diversity via semantic prompts, but still rely on reference images for identity consistency. To enhance expressiveness, multimodal methods incorporate landmarks~\cite{wang2024v}, motion fields~\cite{wang2021audio2head}, or textual cues~\cite{wei2025mocha} to supplement audio. Despite progress, audio remains vulnerable to noise and lacks strong alignment with non-verbal expressions. These limitations motivate a shift toward text-driven paradigms. Our work explores a viseme-aware generation framework that reconstructs expressive lip motion from text alone, bypassing the limitations of unreliable audio input while preserving semantic-visual alignment.

\subsection{Landmark-based Lip Reading}
Lip reading deciphers speech from silent video using visual cues. Early methods rely on handcrafted features~\cite{lucey2008patch, chan2001hmm, luettin1997speechreading}, while deep learning approaches~\cite{stafylakis2017combining} improve spatiotemporal modeling via CNNs. However, most focus on recognition rather than generation. Lip-synchronized face generation reverses the task by synthesizing facial motion from audio. Models like Wav2Lip~\cite{prajwal2020lip}, EAMM~\cite{ji2022eamm}, and SyncTalkFace~\cite{park2022synctalkface} leverage lip landmarks or keypoints as intermediates to align audio and visual dynamics. Despite their success, these methods suffer from phoneme-to-lip ambiguity, where different phonemes share similar lip shapes, limiting generation fidelity. To address this, we introduce visemes—visually discriminative units of speech—as a semantic-visual bridge. By modeling viseme-aware structures, our approach enhances disambiguation and improves expressiveness in text- or audio-driven facial animation.

\section{Methodology}
Although audio-driven talking face generation has made notable progress, it suffers from intrinsic audio-to-lip ambiguity: different phrases (\emph{e.g.}, 'bad boy' vs. 'bat boat') can share nearly identical lip shapes, leading to semantically inconsistent or visually blurred outputs. This ambiguity limits both expressiveness and training stability, especially when semantic precision is needed. To address this, we introduce visemes, visual units that abstract away acoustic differences while capturing shared articulatory patterns, as an intermediate representation. Compared to phonemes, visemes offer a more visually grounded and interpretable signal to guide the movement of the lips. This structured visual prior improves semantic alignment, supports generalization to audio-free settings, and improves generation quality. We therefore propose a viseme-aware pipeline that transforms text into viseme sequences to guide subsequent lip motion synthesis. The detailed process is presented below.

\begin{figure}
  \centering
  \includegraphics[width=\linewidth]{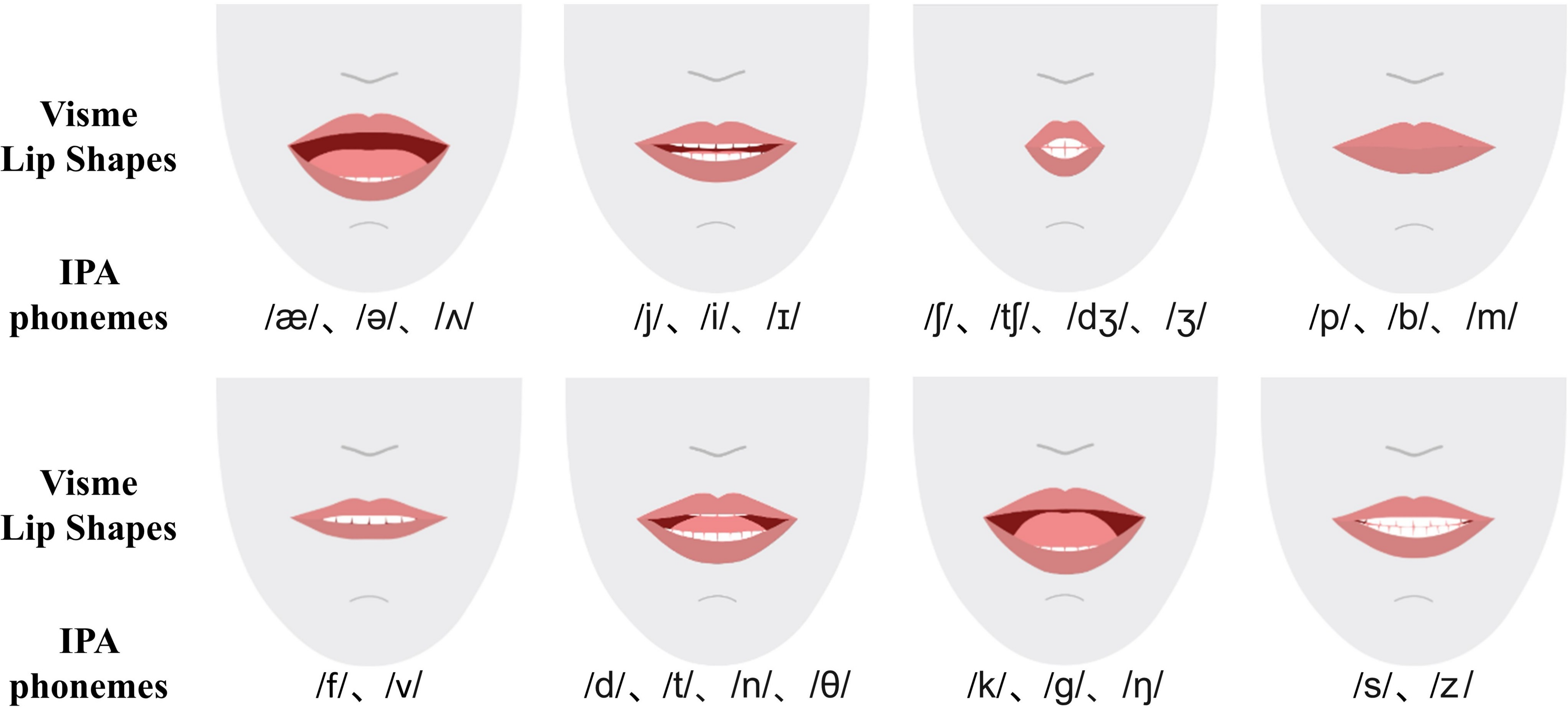}
  \caption{The correspondence between Viseme lip shapes and International Phonetic Alphabet (IPA) phonemes. Each row shows a set of typical lip shapes, and the IPA phonemes that trigger the lip shape are marked below.}
  \label{viseme}
\end{figure}

\subsection{Viseme-Centric Text Encoding}
Accurate lip-reading generation hinges on capturing the visual manifestation of speech articulation. In this context, \emph{visemes}, the minimal visual units representing distinct lip movements, serve as a theoretical and practical bridge between phonetic content and facial motion. Unlike phonemes, visemes abstract away acoustic variations such as vocal cord vibration, and instead focus on articulatory similarities. For instance, while /b/ and /p/ differ phonetically, both correspond to the visual pattern of lips suddenly parting after closure. This visual commonality positions visemes as a crucial intermediate representation for aligning text semantics with lip dynamics.

To visually illustrate the phoneme-to-viseme abstraction, Figure~\ref{viseme} presents a representative mapping between International Phonetic Alphabet (IPA) symbols and their corresponding lip shapes. Each row depicts a typical viseme category, showcasing the shared articulatory appearance among phonemes within the same group. For example, for the phonemes /s/ and /z/, despite differing in vocal cord vibration, exhibit similar tongue and lip configurations, and are thus assigned to the same viseme. In contrast, visually distinct phonemes such as /a:/ and /i/ are grouped into separate viseme classes to preserve critical visual differences.

Building on this insight, we transform input text into fine-grained viseme sequences via a multi-stage pipeline. Specifically, we convert text into IPA sequences using established text-to-phoneme tools, with a dictionary-based fallback for common words to ensure semantic fidelity. Next, a predefined phoneme-to-viseme mapping table (from Microsoft's Speech API) is used to cluster phonemes into viseme categories based on articulatory similarity.

The resulting viseme sequence $\mathbf{V}$ encapsulates the visual articulation structure inherent in the text and serves as an interpretable input for downstream generation. It ensures that lip movements are visually plausible and semantically coherent at the pronunciation level, forming a strong visual prior for accurate and expressive lip-reading synthesis.

\begin{figure*}[th]
  \centering
  \includegraphics[width=\textwidth]{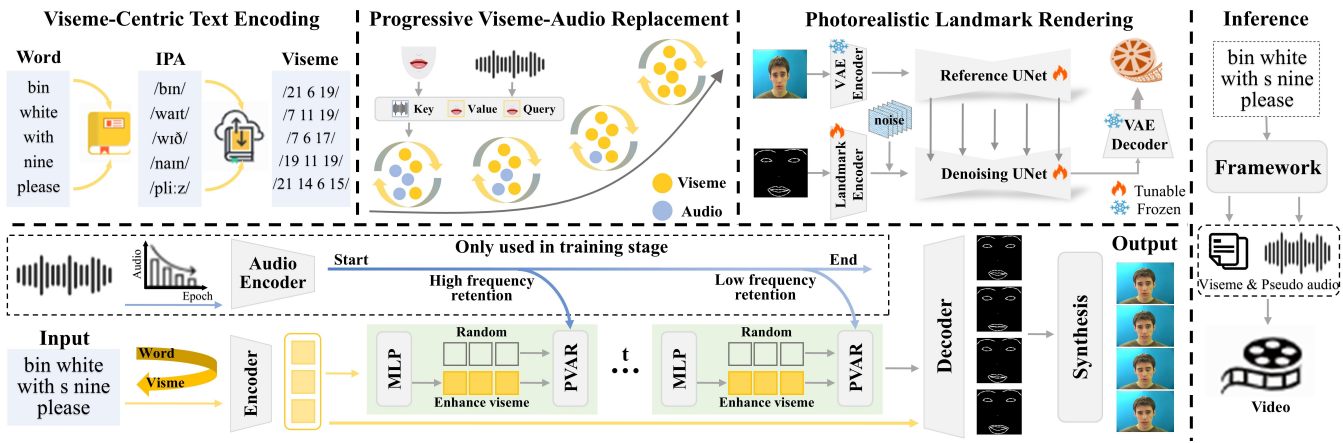}
  \caption{
  Overview of the proposed framework \textbf{Text2Lip}. It contains three key stages: (1) \textit{Viseme-Centric Text Encoding} converts text to visemes via word-phoneme-viseme mapping, (2) \textit{Progressive Viseme-Audio Replacement} adopts text-derived viseme embeddings to radually replace the audio using curriculum learning, and (3) \textit{Photorealistic Landmark Rendering} synthesizes high-quality videos with precise lip synchronization via an adaptive EchoMimic renderer. The inference stage achieves realistic talking face videos driven solely by text.
  }
  \label{main}
\end{figure*}

\subsection{Progressive Viseme-Audio Replacement}
While the viseme sequence $\mathbf{V}$ provides structured and semantically grounded visual instructions, directly generating temporally consistent and expressive facial landmarks from visemes alone remains challenging—especially under real-world conditions where audio is noisy, incomplete, or entirely absent. Most prior methods rely heavily on audio input as the temporal driving force, which limits their robustness and generalizability. 
To address this, we propose \emph{Progressive Viseme-Audio Replacement} (PVAR), a curriculum learning strategy that progressively replaces audio features with viseme-derived from textual embeddings during training. This allows the model to gradually adapt to text-only inputs, while retaining audio guidance in early stages to stabilize training. Furthermore, we introduce a pseudo-audio reconstruction module to recover latent speech information from enhanced text features, ensuring high-fidelity landmark generation even without explicit audio.

\paragraph{Viseme Feature Extraction.}
We employ a Transformer-based encoder to extract high-level semantic features from viseme sequences. Each viseme token is first linearly projected into an embedding space with positional encoding:
\begin{eqnarray}
v'_{n} = W^v \cdot v_n + b^v + PE_v(n),
\end{eqnarray}
where $v_n$ is the one-hot encoding of the $n$-th viseme in vocabulary $\mathcal{V}$, $W^v$ and $b^v$ are learnable parameters, and $PE_v(\cdot)$ denotes sinusoidal position encoding~\cite{vaswani2017transformer}. The resulting embedded sequence $\{v'_n\}_{n=1}^N$ is processed by a viseme encoder:
\begin{eqnarray}
\tilde{v}_{1:N} = \text{VisemeEncoder}(v'_{1:N}).
\end{eqnarray}

\paragraph{Landmark Generation with Modality Replacement.}

To generate accurate facial landmarks under varying modality availability, we design a Transformer-based module that progressively drops the audio stream during training and reconstructs it via pseudo-audio generation. For each time step, facial landmarks and MFCC audio features are encoded as:
\begin{eqnarray}
\left\{
\begin{array}{l}
l'_m = W^l \cdot l_m + b^l + PE_l(m), \\
a'_n = W^a \cdot a_n + b^a + PE_a(n),
\end{array}
\right.
\end{eqnarray}
where $l_m$ is the $m$-th landmark coordinate, $a_n$ is the MFCC audio feature, and $W^l, W^a, b^l, b^a$ are trainable weights and biases.

We introduce an audio dropout mechanism governed by a Bernoulli mask $\mathbf{M}$:
\begin{eqnarray}
\tilde{a}_{1:N} = \mathbf{M} \odot \text{AudioEncoder}(a'_{1:N}), \mathbf{M} \sim B(1 - p_{\text{drop}}),
\end{eqnarray}
where $p_{\text{drop}}$ is linearly increased over training steps:
\begin{eqnarray}
p_{\text{drop}} = p_{\text{start}} + \frac{(p_{\text{end}} - p_{\text{start}}) \cdot t}{T},
\end{eqnarray}
with $t$ denoting the current step and $T$ the total steps. We set $p_{\text{start}}=0$, $p_{\text{end}}=1$ to ensure a full transition to text-only input.

\paragraph{Viseme-Driven Audio Hallucination.}

To compensate for missing audio, we enhance the viseme features using a gated feedforward unit:
\begin{eqnarray}
\tilde{v}^{\text{enh}}_{1:N} = \text{LayerNorm}(\text{GLU}(\tilde{v}_{1:N})).
\end{eqnarray}
These enhanced features are fed into a cross-modal pseudo-audio generator based on multi-head attention:
\begin{eqnarray}
\hat{a}_{1:N} = \text{MultiheadAttention}(\tilde{a}_{1:N}, \tilde{v}^{\text{enh}}_{1:N}, \tilde{v}^{\text{enh}}_{1:N}).
\end{eqnarray}

The reconstructed pseudo-audio $\hat{a}_{1:N}$ is then fused through linear projection and gated modulation:
\begin{eqnarray}
\hat{a}^{\text{pse}}_{1:N} = W_a (\hat{a}_{1:N} \cdot \mathbf{1}_{1 \rightarrow N}) \odot \sigma(\text{Gate}(\hat{a}_{1:N} \cdot \mathbf{1}_{1 \rightarrow N})),
\end{eqnarray}
where $W_a$ is a linear transformation, $\sigma$ is a sigmoid activation, and $\text{Gate}(\cdot)$ denotes a learnable compression layer.

\paragraph{Cross-Modal Landmark Prediction.}

Finally, we use a Transformer-based \textit{LipDecoder} to synthesize facial landmark sequences by integrating pseudo-audio, viseme features, and previously generated poses:
\begin{eqnarray}
\tilde{l}_{m+1} = \text{LipDecoder}(l'_{1:m}, \tilde{v}_{1:N}, \hat{a}^{\text{pse}}_{1:N}),
\end{eqnarray}
which can be decomposed into:
\begin{eqnarray}
\left\{
\begin{array}{l}
z_m = \text{CrossAttention}(l'_{1:m}, \hat{a}^{\text{pse}}_{1:N}, \hat{a}^{\text{pse}}_{1:N}), \\
\tilde{l}_{m+1} = \text{CrossAttention}(z_{1:m}, \tilde{v}_{1:N}, \tilde{v}_{1:N}).
\end{array}
\right.
\end{eqnarray}
This design enables the model to synthesize temporally coherent and semantically aligned facial motion, even under modality-incomplete conditions.

\subsection{Photorealistic Landmark Rendering}

Having obtained temporally coherent and semantically aligned facial landmark sequences $\{\tilde{l}_m\}_{m=1}^{M}$ through PVAR-based compensation, we proceed to synthesize photorealistic talking face videos. This stage translates structural motion cues into pixel-level facial dynamics while preserving lip-audio synchronization and visual realism.

To achieve this, we adopt the state-of-the-art \emph{EchoMimic} model~\cite{chen2025echomimic} as the video rendering backend. EchoMimic is a reference-based video synthesis framework that generates high-fidelity lip-synced talking faces by taking in a sequence of facial landmarks and auxiliary audio features. It employs a spatiotemporal consistency module to enforce smooth lip transitions and a texture completion mechanism to preserve identity and image quality.

Given the predicted landmark trajectory $\{\tilde{l}_{m+1}\}$ and optional audio features (real or pseudo), the video generation process is defined as:
\begin{eqnarray}
\text{Video} = \text{Synthesis}(\{\tilde{l}_{m+1}\}_{m=1}^{M}, \text{audio}),
\end{eqnarray}
where $\text{audio}$ can refer to original audio or reconstructed pseudo-audio, depending on availability. In scenarios where audio input is missing, our framework remains functional by relying solely on viseme-derived pseudo-audio features, ensuring robustness in low-resource or silent input conditions.

This design completes our pipeline from textual input to fully generated talking face video, enabling expressive, semantically accurate, and modality-flexible synthesis.

\subsection{Training Protocol}
We finally streamline our core framework as a three-stage design for text-driven lip-synced talking face generation, as illustrated in Figure~\ref{main}.

\textbf{Stage I: Viseme-Centric Text Encoding} converts input text into a sequence of visual speech units using word-to-phoneme-to-viseme mapping, establishing a semantically aligned prior for lip motion.

\textbf{Stage II: Progressive Viseme-Audio Replacement} employs curriculum learning to gradually replace audio inputs with text-derived viseme embeddings. Cross-modal attention reconstructs pseudo-audio features from enhanced viseme representations, which are fused with visual cues to predict temporally coherent facial landmarks.

\textbf{Stage III: Photorealistic Landmark Rendering} synthesizes high-fidelity videos from landmarks and optional audio (real or pseudo-acoustic) using our adapted EchoMimic renderer, ensuring accurate lip synchronization and temporal smoothness.

This comprehensive framework demonstrates robust performance across different input modalities, supporting both audio-conditioned and text-only generation scenarios while maintaining high visual quality and articulation accuracy. The progressive three-stage design effectively bridges the semantic gap between linguistic inputs and visual outputs through carefully designed intermediate representations.

\begin{table*}[t]
\renewcommand\arraystretch{1.4}
\centering
\resizebox{\textwidth}{!}{
\begin{tabular}{cccc c@{\hskip 0.1pt} ccccccc c@{\hskip 0.1pt} ccccccc}
\Xhline{1pt}
    \multirow{2}{*}{Methods} &\multicolumn{3}{c}{Modality} &~ &\multicolumn{7}{c}{GRID} &~ &\multicolumn{7}{c}{AVDigits}\\
    \cline{2-4}\cline{6-12}\cline{14-20}
    &\textbf{T} &\textbf{A} &\textbf{L} &~ &SSIM{$\uparrow$} &PSNR{$\uparrow$} &LPIPS{$\downarrow$} &FID{$\downarrow$} &FVD{$\downarrow$} &Sync-C{$\uparrow$} &Sync-D{$\downarrow$} &~ &SSIM{$\uparrow$} &PSNR{$\uparrow$} &LPIPS{$\downarrow$} &FID{$\downarrow$} &FVD{$\downarrow$} &Sync-C{$\uparrow$} &Sync-D{$\downarrow$} \\
\Xhline{0.5pt}
    V-Express (arXiv 2024) & &\usym{2714} &\usym{2714} &~ &0.687 &16.693 &0.283 &115.701 &643.881 &4.170 &8.429 &~ &0.719 &15.723 &0.277 &176.767 &562.891 &3.850 &6.445\\ 
    AniPortrait (arXiv 2024) & &\usym{2714} &\usym{2714} &~ &0.698 &17.573 &0.279 &48.722 &452.280 &2.019 &11.907 &~ &0.713 &18.932 &0.252 &88.403 &451.286 &1.349 &9.046\\ 
    EchoMimic (AAAI 2024) & &\usym{2714} &\usym{2714} &~ &0.728 &18.059 &0.274 &38.798 &251.116 &3.687 &8.820 &~ &0.728 &19.128 &0.241 &61.237 &279.188 &3.692 &6.764\\ 
    Hallo2 (ICLR 2025) &\usym{2714} &\usym{2714} & &~ &0.733 &17.532 &0.291 &69.896 &287.288 &4.170 &8.299 &~ &0.735 &19.355 &0.247 &57.625 &311.571 &3.694 &6.490\\ 
    SadTalker (CVPR 2023) & &\usym{2714} & &~ &0.721 &18.372 &0.297 &195.381 &341.328 &5.661 &7.403 &~ &0.717 &18.663 &0.250 &154.211 &308.519 &3.198 &{\bfseries 5.742}\\
    Sonic (CVPR 2025) & &\usym{2714} & &~ &0.737 &18.821 &0.241 &34.128 &458.257 &{\bfseries 5.851} &{\bfseries 7.038} &~ &0.739 &19.621 &0.240 &55.118 &266.924 &{\bfseries 3.892} &6.317\\ 
\Xhline{0.5pt}
    Text2Lip (Ours) &\usym{2714} & & &~ &{\bfseries 0.740} &{\bfseries 19.023} &{\bfseries 0.238} &{\bfseries 32.109} &{\bfseries 277.655} &4.641 &7.302 &~ &{\bfseries 0.741} &{\bfseries 19.721} &{\bfseries 0.238} &{\bfseries 51.282} &{\bfseries 251.441} &3.715 &6.207\\
\Xhline{1pt}
\end{tabular}}
\caption{Quantitative comparisons with the state-of-the-arts on GRID and AVDigits datasets. \usym{2714} indicates the modality used. \textbf{T}, 
\textbf{A} and \textbf{L} represent Text, Audio and Landmark respectively.}
\label{Quantitative comparisons with the state-of-the-arts on GRID and AVDigits datasets.}
\end{table*}

\begin{table}[t]
\renewcommand\arraystretch{1.4}
\centering
\resizebox{0.48\textwidth}{!}{
\begin{tabular}{cccccccc}
\Xhline{1pt}
   Methods &BLEU-1{$\uparrow$} &BLEU-4{$\uparrow$} &WER{$\downarrow$} &DTW-P{$\downarrow$} &MPJPE{$\downarrow$}\\
\Xhline{0.5pt}
PT-GN (ECCV 2020)  &40.87 &9.76 &52.17 &4.97 &807.49\\
GEN-OBT (MM 2022)  &36.36 &6.35 &56.33 &6.55 &{\bfseries 761.35}\\
LVMCN (ICASSP 2025)  &47.85 &15.36 &45.41 &5.32 &807.37\\
\Xhline{0.5pt}
Text2Lip(Ours) &{\bfseries 54.81} &{\bfseries 23.50} &{\bfseries 39.43} &{\bfseries 3.27} &783.32\\
\Xhline{1pt}
\end{tabular}}
\caption{Landmark comparison results on GRID dataset.}
\label{Landmark comparison results on GRID dataset.}
\end{table}

\section{Experiments}
\subsection{Experimental Settings}
\subsubsection{Datasets.} 
Following existing works~\cite{assael2016lipnet,wang2021audio2head}, we conduct experiments on two popular benchmarks \textbf{GRID}~\cite{cooke2006audio} and \textbf{AVDigits}~\cite{hu2016temporal} to evaluate the effectiveness of the proposed method.
GRID contains videos of 33 speakers, covering a total of 51 words. Each sentence consists of six words, with a fixed sentence structure of command + color + preposition + letter + number + adverb (\emph{e.g.}, lay red with y two again). All videos have a consistent duration of 75 frames. AVDigits captures the speech video of digits 0 to 9 spoken nine times by six speakers. Each video is recorded at 25 fps and the audio is recorded at 48 kHz.
To improve data quality, we screen the original videos, removing abnormal samples such as those with excessive head movements, multiple faces, and inconsistencies in the speaker's voice. The processed data are divided into training, validation, and test sets according to ~\cite{assael2016lipnet}. In addition, we also test our model on self-collected data for open-domain evaluation.

\subsubsection{Evaluation Metrics.}
We use multiple indicators to perform quantitative analysis from different dimensions. First, we use the \textbf{SSIM}~\cite{wang2004image},  \textbf{PSNR}~\cite{hore2010image}, \textbf{LPIPS}~\cite{zhang2018unreasonable}, \textbf{FID}~\cite{heusel2017gans} and \textbf{FVD}~\cite{unterthiner2018towards} to measure the visual similarity between the generated videos and the real videos at the image and video levels. \textbf{DTW-P}~\cite{sakoe2003dynamic} and \textbf{MPJPE}~\cite{ionescu2013human3} are adopted to measure the matching degree between the generated landmark sequence with ground-truth. 
Besides, we introduce the SyncNet~\cite{Chung16a} model to calculate \textbf{Sync-C} and \textbf{Sync-D}, in which Sync-C measures the synchronization consistency between lip sounds and audio, while the Sync-D evaluates the temporal consistency of dynamic lip movements. 
In addition, we design an inverse evaluation to measure the semantic preservation accuracy of generated videos. Specifically, we retrain the NSLT~\cite{camgoz2018neural} model, which interprets text sentences from lip reading videos and calculates the text alignment accuracy, such as \textbf{BLEU}~\cite{papineni2002bleu}, \textbf{ROUGE}~\cite{lin2004rouge}, \textbf{WER}~\cite{assael2016lipnet}, \textbf{DTW-P}~\cite{sakoe2003dynamic}.

\subsubsection{Data Prepossessing.} 
We adopt the DLib~\cite{kazemi2014one} library to accurately extract the 2D coordinates of 68 facial landmarks from the original videos.
The extracted coordinates are globally normalized to eliminate individual and environmental variations, producing standardized 2D facial landmark labels. For audio data, we use the Mel-frequency cepstral coefficients (MFCC) to preprocess the original audio data, laying a foundation for subsequent model training and analysis.

\subsubsection{Model Settings.}
Our Text2Lip model is built from 2-layer, 4-head Transformers with a uniform embedding size of 512. All parts of the network are trained using the Adam optimizer with a learning rate of $1 \times {10}^{-3}$ and a batch size of \textit{batchsize} = 128. All experiments are conducted on a server equipped with 8 NVIDIA A40 GPUs.

\subsection{Comparison with State-of-the-Arts}
We compare the Text2Lip with some mainstream talkface generation methods: V-Express~\cite{wang2024v}, AniPortrait~\cite{wei2024aniportrait}, EchoMimic~\cite{chen2025echomimic}, Hallo2~\cite{cuihallo2}, SadTalker~\cite{zhang2023sadtalker} and Sonic~\cite{ji2025sonic}. It is worth noting that all compared methods rely on audio input, and some of them incorporate additional signals such as text or landmarks, while our model can be deployed under audio-free conditions.
    
\subsubsection{Comparison on GRID.} As shown in Table~\ref{Quantitative comparisons with the state-of-the-arts on GRID and AVDigits datasets.}, our method Text2Lip achieves superior or comparable performance across all major evaluation metrics, demonstrating the effectiveness of our audio-free design. In terms of visual quality, our model achieves the highest SSIM of 0.740 and PSNR of 19.023, demonstrating superior structural similarity and pixel-level fidelity compared to methods leveraging audio (\emph{e.g.}, Sonic: SSIM 0.737, PSNR 18.821). Furthermore, we achieve the lowest LPIPS of 0.238 and FID of 32.109, highlighting enhanced perceptual realism and distributional closeness to real videos.
In terms of overall video quality, our method achieves the best FVD score of 277.65, outperforming all compared methods, including those utilizing audio and keypoints. While lip sync accuracy is inherently challenging in the absence of audio, Text2Lip still achieves Sync-C 4.641 and Sync-D 7.302, surpassing V-Express and EchoMimic and approaching the performance of strong audio-based methods SadTalker and Sonic. Despite relying solely on text, Text2Lip surpasses or approaches the audio-based baseline in all metrics, demonstrating the feasibility and advantages of audio-free speaking face generation in real-world scenarios.

We further evaluate Text2Lip landmark generation capabilities on GRID dataset, with the results shown in Table~\ref{Landmark comparison results on GRID dataset.}. Text2Lip achieve a BLEU-1 score of 54.81 and a BLEU-4 score of 23.50, significantly outperforming state-of-the-art methods such as PT-GN~\cite{saunders2020ptslp}, GEN-OBT~\cite{tang2022gloss}, and LVMCN~\cite{wang2025linguistics}, demonstrating higher semantic fidelity in landmark sequences. Its WER of 39.43 is lower than all other compared methods, further confirming its accurate landmark sequence reconstruction. In terms of dynamic metrics, although its MPJPE 783.32 performance is lower than GEN-OBT, overall, we achieve a balance between semantics and landmark accuracy.

\subsubsection{Comparison on AVDigits.}
As shown in Table~\ref{Quantitative comparisons with the state-of-the-arts on GRID and AVDigits datasets.}, Text2Lip also demonstrates highly competitive performance on the AVDigits dataset. Despite relying solely on text, it achieves a FVD score of 251.841, surpassing strong baselines such as V-Express (176.67), Sonic (308.594), and SadTalker (451.286). In terms of lip synchronization, Text2Lip achieves Sync-C scores of 3.715 and Sync-D scores of 6.207, comparable to or exceeding audio-based methods such as SadTalker (3.198/6.317) and V-Express (3.850/6.445), demonstrating its ability to ensure lip sync coherence without audio supervision. These results validate the feasibility of the Text2Lip text-only framework for generating natural, synchronized speaking faces and extend its utility to scenarios with limited or no audio.

\subsubsection{User Study.}
We conduct subjective evaluations on GRID and AVDigits datasets  across four key dimensions: naturalness, visual clarity, temporal consistency, and smoothness. A total of 100 participants rated the results of the five compared methods on a scale of 1 to 5. As shown in Table~\ref{User study comparison on GRID dataset.}, our Text2Lip method outperformed the other methods across all four dimensions, particularly achieving a significant 37\% improvement in smoothness.

\begin{table}[t]
\renewcommand\arraystretch{1.4}
\centering
\resizebox{0.48\textwidth}{!}{
\begin{tabular}{c ccc c@{\hskip 0.1pt} cccc}
\Xhline{1pt}
    \multirow{2}{*}{Sources} &\multicolumn{3}{c}{Semantic Quality} &~ &\multicolumn{4}{c}{Video Quality}\\
    \cline{2-4}\cline{6-9}   
     &BLEU-1{$\uparrow$} &BLEU-4{$\downarrow$} &WER{$\downarrow$} &~ &SSIM{$\uparrow$} &PSNR{$\uparrow$} &LPIPS{$\downarrow$} &FID{$\downarrow$}\\
\Xhline{0.5pt}
GT &55.20 &24.45 &39.48 &~ &0.743 &19.217 &0.223 &30.296 \\
\Xhline{0.5pt}
From IndexTTS &49.46 &16.82 &44.29 &~ &0.718 &17.631 &0.281 &91.276\\
From offline coding &51.86 &19.72 &41.84 &~ &0.721 &17.859 &0.276 &87.655\\
From viseme (Ours) &{\bfseries 54.81} &{\bfseries 23.50} &{\bfseries 39.43} &~ &{\bfseries 0.740} &{\bfseries 19.023} &{\bfseries 0.238} &{\bfseries 32.109}\\
\Xhline{1pt}
\end{tabular}}
\caption{Comparison results of different audio sources on GRID dataset.}
\label{Comparison results of audio from different channels on GRID dataset.}
\end{table}

\begin{table}[t]
\renewcommand\arraystretch{1.4}
\centering
\resizebox{0.48\textwidth}{!}{
\begin{tabular}{c ccc c@{\hskip 0.1pt} cccc}
\Xhline{1pt}
    \multirow{2}{*}{Methods} &\multicolumn{3}{c}{Semantic Quality} &~ &\multicolumn{4}{c}{Video Quality}\\
    \cline{2-4}\cline{6-9}   
     &BLEU-1{$\uparrow$} &BLEU-4{$\downarrow$} &WER{$\downarrow$} &~ &SSIM{$\uparrow$} &PSNR{$\uparrow$} &LPIPS{$\downarrow$} &FID{$\downarrow$}\\
\Xhline{0.5pt}
    Text &40.04 &9.91 &52.64 &~ &0.681 &15.896 &0.301 &187.628\\
    Text $\rightarrow$ Viseme &47.23 &14.78 &45.76 &~ &0.702 &17.553 &0.292 &120.391\\
    Text2Lip (Ours) &{\bfseries 54.81} &{\bfseries 23.50} &{\bfseries 39.43} &~ &{\bfseries 0.740} &{\bfseries 19.023} &{\bfseries 0.238} &{\bfseries 32.109}\\
\Xhline{1pt}
\end{tabular}}
\caption{The role of viseme conversion strategy and pseudo-audio generation module on GRID dataset.}
\label{The role of viseme conversion strategy and pseudo-audio generation module on GRID dataset.}
\end{table}

\begin{table}[t]

\renewcommand\arraystretch{1.4}
\centering
\resizebox{0.48\textwidth}{!}{
\begin{tabular}{ccccc}
\Xhline{1pt}
     Methods &Naturalness &Visual clarity &Temporal consistency  &Smoothness \\
\Xhline{0.5pt}
    Sonic &3.35 &3.16 &3.22 &3.05\\
    Hallo2 &2.82 &2.92 &2.71 &2.75\\
    AniPortrait &2.57 &3.01 &2.63 &2.71\\
    V-Express &1.62 &2.19 &2.31 &2.28\\
\Xhline{0.5pt}
    Text2Lip (Ours) &{\bfseries 3.97 (19\%$\uparrow$)} &{\bfseries 4.23 (34\%$\uparrow$)} &{\bfseries 4.26 (32\%$\uparrow$)} &{\bfseries 4.19 (37\%$\uparrow$)}\\
\Xhline{1pt}
\end{tabular}}
\caption{User study comparison on GRID dataset.}
\label{User study comparison on GRID dataset.}
\end{table}

\subsection{Ablation Study}
\subsubsection{The Role of Viseme-centric Text Encoding.}
The experimental analysis in Table~\ref{The role of viseme conversion strategy and pseudo-audio generation module on GRID dataset.} shows that viseme-centric encoding effectively bridges the semantic gap, resolves phoneme ambiguity, and significantly improves semantic quality indicators (BLEU-1 increases by 17.9\% to 47.23, and WER decreases by 13.1\% to 45.76). It also enhances lip movement coherence and promotes video quality optimization (FID decreases to 120.391, and SSIM increases by 0.702).
Furthermore, the PVAR strategy gradually replaces audio features with text embeddings during training through curriculum learning, and collaborates with the pseudo-audio generation module to re-encode the audio features. The new method uses acoustic constraints to achieve a performance leap in the absence of audio: at the semantic level, BLEU-4 significantly increases by 59.1\% to 23.50, and WER further decreases by 13.8\% to 39.43; at the visual fidelity level, FID plummets by 73.3\% to 32.109, LPIPS increases by 18.5\% to 0.238, and PSNR increases by 8.4\% to 19.023, confirming the reasonable guidance of pseudo-audio on lip movements and verifying that Text2Lip provides an explainable technical paradigm for cross-modal generation.

\subsubsection{Effect of Progressive Viseme-Audio Replacement.}
As shown in Table~\ref{Comparison results of audio from different channels on GRID dataset.}, we compare three audio sources: audio generated by IndexTTS, offline coding audio, and our proposed pseudo-audio. Our method achieves superior semantic quality (BLEU-1: 54.81, BLEU-4: 23.50, WER: 39.43) compared to both IndexTTS and offline encoding. In terms of visual quality, our method also surpasses the baselines in SSIM (0.740), PSNR (19.023), and FID (32.109), demonstrating improvements in both semantic accuracy and visual realism. Notably, our pseudo-audio achieves performance close to that of ground-truth audio, demonstrating its ability to effectively capture the speech features crucial for high-quality speaking face generation and effectively achieving the desired effect of audio replacement.

\begin{figure}[t]
  \centering
  \includegraphics[width=\linewidth]{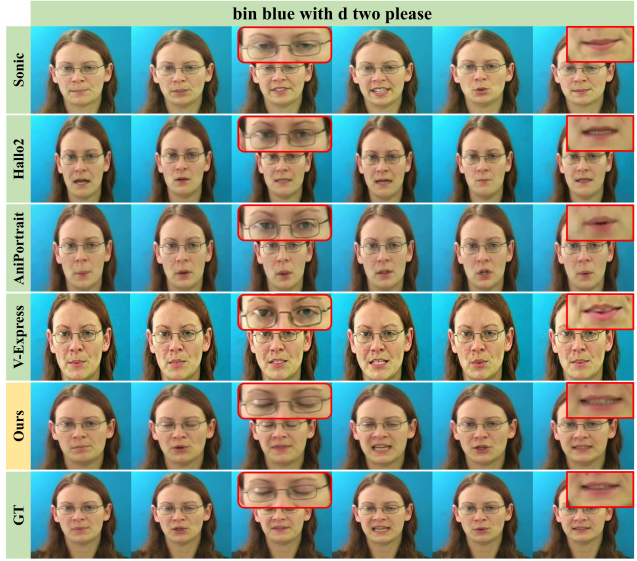}
  \caption{
  Visualization examples on GRID dataset.
  }
  \label{Visualization examples on GRID dataset.}
\end{figure}

\begin{figure}[t]
  \centering
  \includegraphics[width=\linewidth]{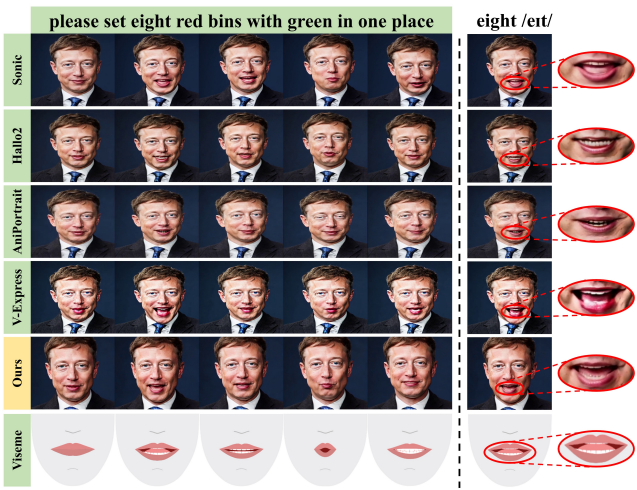}
  \caption{Visualization examples of unseen sentences in open domains.
  }
  \label{Visualization examples of unseen sentences in the open domain.}
\end{figure}

\subsection{Qualitative Results}
\subsubsection{Visualization Results on dataset.} Figure~\ref{Visualization examples on GRID dataset.} shows a qualitative comparison between baseline methods (Sonic, Hallo2, AniPortrait and V-Express) and our method on GRID dataset, focusing on the third column (blink dynamics) and the sixth column (lip shape details) with significant differences.
In terms of tone-driven blinks, the compared methods suffer from insufficient closure (\emph{e.g.}, residual eyelid occlusion in Sonic and Hallo2) and abrupt transitions (\emph{e.g.}, uniform occlusion lacking gradualness in AniPortrait). Our method accurately captures eyelid edge alignment and light-shadow transitions, and its closure duration and opening and closing amplitude are highly consistent with the GT.
Regarding lip shape details, the comparisons exhibit blurred contours (\emph{e.g.}, jagged edges in V-Express) and motion asymmetry (\emph{e.g.}, lip deviation in AniPortrait). Our method generates lip shapes with clear anatomical structure, a natural transition between the upper lip peak and lower lip valley that aligns with articulatory dynamics, and a closer approximation to the GT.
The results show that Text2Lip can achieve fine coordination of eyebrow movement, eyelid opening and closing, and lip shape while keeping the head stable, and is overall closer to the natural human expressions.

\subsubsection{Performance on Unseen Sentences.}
To verify its performance on unseen open-domain sentences, we use the real complex sentence “please set eight red bins with green in one place” for visualization analysis.
As shown in Figure~\ref{Visualization examples of unseen sentences in the open domain.}, compared to mainstream methods such as Sonic, Hallo2, AniPortrait and V-Expres, our method significantly outperforms in lip shape naturalness, lip fluency, and semantic relevance: The lip shape space closely matches the articulatory movements (\emph{e.g.}, lips open in “eight” and lips closed in “set”), and generalization is stable in the open domain (even robust to untrained sentences).
The enlarged image further verifies that key phonemes (\emph{e.g.}, lips open in the vowel “eight”  \textipa{/eI/} and lips and teeth contact in the final /t/) are strictly mapped to viseme lip shape, demonstrating the advantages of viseme decoupled representation. This method enables precise control of lip-synced talking faces generation in complex open-domain scenarios, providing technical support for cross-modal visual interaction.

\begin{figure}[t]
  \centering
  \includegraphics[width=\linewidth]{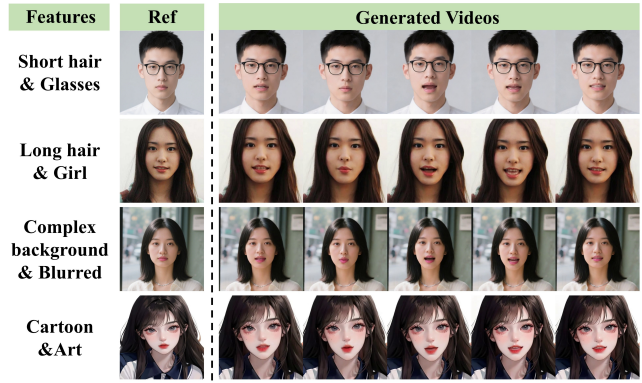}
  \caption{Visualization examples of unseen speakers in open domains.
  }
  \label{Visualization examples of unseen speakers in the open domain.}
\end{figure}

\subsubsection{Performance on Unseen Speakers.}
As shown in the visualization in Figure~\ref{Visualization examples of unseen speakers in the open domain.}, we conducted validation analysis on unseen speakers in an open-domain scenario. For character settings with differentiated basic appearance features, such as boys with glasses and long-haired girls, the generated video sequences accurately reproduced the speech-synchronized lip movements and micro-expression dynamics of the reference image while preserving key features of the target identity. Furthermore, in complex, blurred scenes, the algorithm effectively separated foreground characters from interfering background information. The generated dynamic images preserved the subject's facial details while also appropriately filling in the blurred areas through adaptive background generation, verifying the model's stability under the influence of environmental variables. In the challenging cartoon art style task, the generated results successfully captured the line features and color semantics of the hand-drawn style, transforming the real-life reference into a virtual avatar with a distinctive artistic style. The overall results show that the identity-separated lip-sync video generation strategy can be flexibly transferred to different roles and scenarios, providing an effective solution for speaker-unseen talking face generation in open domains.

\section{Conclusions}
In this work, we propose a viseme-centric framework, Text2Lip, that bridges linguistic semantics and visual articulation for viseme-driven lip-synced talking face generation. The proposed method addresses key limitations of conventional audio-driven approaches through three core innovations: structured viseme mapping for semantic-aligned lip motion priors, curriculum-based viseme-audio replacement for robust multi-modal handling, and landmark-guided rendering for photorealistic synthesis. Extensive evaluations demonstrate superior performance in both semantic consistency, visual quality, and generalization in open domains.

\bibliography{references}

\begin{thebibliography}{46}
\providecommand{\natexlab}[1]{#1}

\bibitem[{Assael et~al.(2016)Assael, Shillingford, Whiteson, and De~Freitas}]{assael2016lipnet}
Assael, Y.~M.; Shillingford, B.; Whiteson, S.; and De~Freitas, N. 2016.
\newblock Lipnet: End-to-end sentence-level lipreading.
\newblock \emph{arXiv preprint arXiv:1611.01599}.

\bibitem[{Camgoz et~al.(2018)Camgoz, Hadfield, Koller, Ney, and Bowden}]{camgoz2018neural}
Camgoz, N.~C.; Hadfield, S.; Koller, O.; Ney, H.; and Bowden, R. 2018.
\newblock Neural Sign Language Translation.
\newblock In \emph{Computer Vision and Pattern Recognition}, 7784--7793.

\bibitem[{Chan(2001)}]{chan2001hmm}
Chan, M.~T. 2001.
\newblock HMM-based audio-visual speech recognition integrating geometric-and appearance-based visual features.
\newblock In \emph{IEEE Fourth Workshop on Multimedia Signal Processing}, 9--14.

\bibitem[{Chen et~al.(2025)Chen, Cao, Chen, Li, and Ma}]{chen2025echomimic}
Chen, Z.; Cao, J.; Chen, Z.; Li, Y.; and Ma, C. 2025.
\newblock Echomimic: Lifelike audio-driven portrait animations through editable landmark conditions.
\newblock In \emph{Proceedings of the AAAI Conference on Artificial Intelligence}, 2403--2410.

\bibitem[{Chung and Zisserman(2016)}]{Chung16a}
Chung, J.~S.; and Zisserman, A. 2016.
\newblock Out of time: automated lip sync in the wild.
\newblock In \emph{Workshop on Multi-view Lip-reading, ACCV}.

\bibitem[{Cooke et~al.(2006)Cooke, Barker, Cunningham, and Shao}]{cooke2006audio}
Cooke, M.; Barker, J.; Cunningham, S.; and Shao, X. 2006.
\newblock An audio-visual corpus for speech perception and automatic speech recognition.
\newblock \emph{The Journal of the Acoustical Society of America}, 120(5): 2421--2424.

\bibitem[{Cui et~al.(2025{\natexlab{a}})Cui, Li, Yao, Zhu, Shang, Cheng, Zhou, Zhu, and Wang}]{cuihallo2}
Cui, J.; Li, H.; Yao, Y.; Zhu, H.; Shang, H.; Cheng, K.; Zhou, H.; Zhu, S.; and Wang, J. 2025{\natexlab{a}}.
\newblock Hallo2: Long-Duration and High-Resolution Audio-Driven Portrait Image Animation.
\newblock In \emph{The Thirteenth International Conference on Learning Representations}.

\bibitem[{Cui et~al.(2025{\natexlab{b}})Cui, Li, Zhan, Shang, Cheng, Ma, Mu, Zhou, Wang, and Zhu}]{cui2025hallo3}
Cui, J.; Li, H.; Zhan, Y.; Shang, H.; Cheng, K.; Ma, Y.; Mu, S.; Zhou, H.; Wang, J.; and Zhu, S. 2025{\natexlab{b}}.
\newblock Hallo3: Highly dynamic and realistic portrait image animation with video diffusion transformer.
\newblock In \emph{Proceedings of the Computer Vision and Pattern Recognition Conference}, 21086--21095.

\bibitem[{Ferdowsifard et~al.(2021)Ferdowsifard, Barke, Peleg, Lerner, and Polikarpova}]{ferdowsifard2021loopy}
Ferdowsifard, K.; Barke, S.; Peleg, H.; Lerner, S.; and Polikarpova, N. 2021.
\newblock LooPy: interactive program synthesis with control structures.
\newblock \emph{Proceedings of the ACM on Programming Languages}, 5(OOPSLA): 1--29.

\bibitem[{Guo, Tang, and Wang(2019)}]{Guo_Tang}
Guo, D.; Tang, S.; and Wang, M. 2019.
\newblock Connectionist Temporal Modeling of Video and Language: A Joint Model for Translation and Sign Labeling.
\newblock In \emph{International Joint Conference on Artificial Intelligence}, 751--757.

\bibitem[{Heusel et~al.(2017)Heusel, Ramsauer, Unterthiner, Nessler, and Hochreiter}]{heusel2017gans}
Heusel, M.; Ramsauer, H.; Unterthiner, T.; Nessler, B.; and Hochreiter, S. 2017.
\newblock Gans trained by a two time-scale update rule converge to a local nash equilibrium.
\newblock \emph{Advances in neural information processing systems}, 30.

\bibitem[{Hore and Ziou(2010)}]{hore2010image}
Hore, A.; and Ziou, D. 2010.
\newblock Image quality metrics: PSNR vs. SSIM.
\newblock In \emph{2010 20th international conference on pattern recognition}, 2366--2369. IEEE.

\bibitem[{Hu, Li et~al.(2016)}]{hu2016temporal}
Hu, D.; Li, X.; et~al. 2016.
\newblock Temporal multimodal learning in audiovisual speech recognition.
\newblock In \emph{Proceedings of the IEEE Conference on Computer Vision and Pattern Recognition}, 3574--3582.

\bibitem[{Ionescu et~al.(2013)Ionescu, Papava, Olaru, and Sminchisescu}]{ionescu2013human3}
Ionescu, C.; Papava, D.; Olaru, V.; and Sminchisescu, C. 2013.
\newblock Human3. 6m: Large scale datasets and predictive methods for 3d human sensing in natural environments.
\newblock \emph{IEEE transactions on pattern analysis and machine intelligence}, 36(7): 1325--1339.

\bibitem[{Ji et~al.(2025)Ji, Hu, Xu, Zhu, Lin, He, Zhang, Luo, Chen, Lin et~al.}]{ji2025sonic}
Ji, X.; Hu, X.; Xu, Z.; Zhu, J.; Lin, C.; He, Q.; Zhang, J.; Luo, D.; Chen, Y.; Lin, Q.; et~al. 2025.
\newblock Sonic: Shifting focus to global audio perception in portrait animation.
\newblock In \emph{Proceedings of the Computer Vision and Pattern Recognition Conference}, 193--203.

\bibitem[{Ji et~al.(2022)Ji, Zhou, Wang, Wu, Wu, Xu, and Cao}]{ji2022eamm}
Ji, X.; Zhou, H.; Wang, K.; Wu, Q.; Wu, W.; Xu, F.; and Cao, X. 2022.
\newblock Eamm: One-shot emotional talking face via audio-based emotion-aware motion model.
\newblock In \emph{ACM SIGGRAPH 2022 Conference Proceedings}, 1--10.

\bibitem[{Kazemi and Sullivan(2014)}]{kazemi2014one}
Kazemi, V.; and Sullivan, J. 2014.
\newblock One millisecond face alignment with an ensemble of regression trees.
\newblock In \emph{Proceedings of the IEEE conference on computer vision and pattern recognition}, 1867--1874.

\bibitem[{Lin(2004)}]{lin2004rouge}
Lin, C.-Y. 2004.
\newblock Rouge: A package for automatic evaluation of summaries.
\newblock In \emph{Text summarization branches out}, 74--81.

\bibitem[{Lucey, Potamianos, and Sridharan(2008)}]{lucey2008patch}
Lucey, P.; Potamianos, G.; and Sridharan, S. 2008.
\newblock Patch-based analysis of visual speech from multiple views.
\newblock In \emph{International Conference on Auditory-Visual Speech Processing}, 69--74.

\bibitem[{Luettin and Thacker(1997)}]{luettin1997speechreading}
Luettin, J.; and Thacker, N.~A. 1997.
\newblock Speechreading using probabilistic models.
\newblock \emph{Computer vision and image understanding}, 65(2): 163--178.

\bibitem[{Papineni et~al.(2002)Papineni, Roukos, Ward, and Zhu}]{papineni2002bleu}
Papineni, K.; Roukos, S.; Ward, T.; and Zhu, W.-J. 2002.
\newblock Bleu: a method for automatic evaluation of machine translation.
\newblock In \emph{Proceedings of the 40th annual meeting of the Association for Computational Linguistics}, 311--318.

\bibitem[{Park et~al.(2022)Park, Kim, Hong, Choi, and Ro}]{park2022synctalkface}
Park, S.~J.; Kim, M.; Hong, J.; Choi, J.; and Ro, Y.~M. 2022.
\newblock Synctalkface: Talking face generation with precise lip-syncing via audio-lip memory.
\newblock In \emph{AAAI Conference on Artificial Intelligence}, 2062--2070.

\bibitem[{Prajwal et~al.(2020)Prajwal, Mukhopadhyay, Namboodiri, and Jawahar}]{prajwal2020lip}
Prajwal, K.; Mukhopadhyay, R.; Namboodiri, V.~P.; and Jawahar, C. 2020.
\newblock A lip sync expert is all you need for speech to lip generation in the wild.
\newblock In \emph{ACM international conference on multimedia}, 484--492.

\bibitem[{Sakoe and Chiba(2003)}]{sakoe2003dynamic}
Sakoe, H.; and Chiba, S. 2003.
\newblock Dynamic programming algorithm optimization for spoken word recognition.
\newblock \emph{IEEE transactions on acoustics, speech, and signal processing}, 26(1): 43--49.

\bibitem[{Saunders, Camgoz, and Bowden(2020)}]{saunders2020ptslp}
Saunders, B.; Camgoz, N.~C.; and Bowden, R. 2020.
\newblock Progressive Transformers for End-to-End Sign Language Production.
\newblock In \emph{European Conference on Computer Vision}, 687--705.

\bibitem[{Stafylakis and Tzimiropoulos(2017)}]{stafylakis2017combining}
Stafylakis, T.; and Tzimiropoulos, G. 2017.
\newblock Combining residual networks with LSTMs for lipreading.
\newblock \emph{arXiv preprint arXiv:1703.04105}.

\bibitem[{Tang et~al.(2022{\natexlab{a}})Tang, Guo, Hong, and Wang}]{tangslt2022}
Tang, S.; Guo, D.; Hong, R.; and Wang, M. 2022{\natexlab{a}}.
\newblock Graph-Based Multimodal Sequential Embedding for Sign Language Translation.
\newblock \emph{IEEE Transactions on Multimedia}, 4433--4445.

\bibitem[{Tang et~al.(2025{\natexlab{a}})Tang, He, Cheng, Wu, Guo, and Hong}]{tang2025discrete}
Tang, S.; He, J.; Cheng, L.; Wu, J.; Guo, D.; and Hong, R. 2025{\natexlab{a}}.
\newblock Discrete to Continuous: Generating Smooth Transition Poses from Sign Language Observations.
\newblock In \emph{Computer Vision and Pattern Recognition Conference}, 3481--3491.

\bibitem[{Tang et~al.(2025{\natexlab{b}})Tang, He, Guo, Wei, Li, and Hong}]{tang2025sign}
Tang, S.; He, J.; Guo, D.; Wei, Y.; Li, F.; and Hong, R. 2025{\natexlab{b}}.
\newblock Sign-IDD: Iconicity Disentangled Diffusion for Sign Language Production.
\newblock In \emph{AAAI Conference on Artificial Intelligence}, 7266--7274.

\bibitem[{Tang et~al.(2022{\natexlab{b}})Tang, Hong, Guo, and Wang}]{tang2022gloss}
Tang, S.; Hong, R.; Guo, D.; and Wang, M. 2022{\natexlab{b}}.
\newblock Gloss Semantic-Enhanced Network with Online Back-Translation for Sign Language Production.
\newblock In \emph{ACM International Conference on Multimedia}, 5630--5638.

\bibitem[{Tang et~al.(2025{\natexlab{c}})Tang, Xue, Wu, Wang, and Hong}]{tang2024GCDM}
Tang, S.; Xue, F.; Wu, J.; Wang, S.; and Hong, R. 2025{\natexlab{c}}.
\newblock Gloss-Driven Conditional Diffusion Models for Sign Language Production.
\newblock \emph{ACM Transactions on Multimedia Computing, Communications and Applications}, 1--17.

\bibitem[{Unterthiner et~al.(2018)Unterthiner, Van~Steenkiste, Kurach, Marinier, Michalski, and Gelly}]{unterthiner2018towards}
Unterthiner, T.; Van~Steenkiste, S.; Kurach, K.; Marinier, R.; Michalski, M.; and Gelly, S. 2018.
\newblock Towards accurate generative models of video: A new metric \& challenges.
\newblock \emph{arXiv preprint arXiv:1812.01717}.

\bibitem[{Vaswani et~al.(2017)Vaswani, Shazeer, Parmar, Uszkoreit, Jones, Gomez, Kaiser, and Polosukhin}]{vaswani2017transformer}
Vaswani, A.; Shazeer, N.; Parmar, N.; Uszkoreit, J.; Jones, L.; Gomez, A.~N.; Kaiser, {\L}.; and Polosukhin, I. 2017.
\newblock Attention is All You Need.
\newblock In \emph{Neural Information Processing Systems}, 1--11.

\bibitem[{Wang et~al.(2024)Wang, Tian, Zhang, Guan, Luo, Shen, Jiang, Gu, Han, and Yang}]{wang2024v}
Wang, C.; Tian, K.; Zhang, J.; Guan, Y.; Luo, F.; Shen, F.; Jiang, Z.; Gu, Q.; Han, X.; and Yang, W. 2024.
\newblock V-express: Conditional dropout for progressive training of portrait video generation.
\newblock \emph{arXiv preprint arXiv:2406.02511}.

\bibitem[{Wang et~al.(2021)Wang, Li, Ding, Fan, and Yu}]{wang2021audio2head}
Wang, S.; Li, L.; Ding, Y.; Fan, C.; and Yu, X. 2021.
\newblock Audio2Head: Audio-driven One-shot Talking-head Generation with Natural Head Motion.
\newblock In \emph{Proceedings of the Thirtieth International Joint Conference On Artificial Intelligence, Ijcai 2021}, 1098--1105. International Joint Conferences on Artificial Intelligence Organization.

\bibitem[{Wang et~al.(2025{\natexlab{a}})Wang, Tang, Cheng, Li, Wang, and Hong}]{wang2025signaligner}
Wang, X.; Tang, S.; Cheng, L.; Li, F.; Wang, S.; and Hong, R. 2025{\natexlab{a}}.
\newblock SignAligner: Harmonizing Complementary Pose Modalities for Coherent Sign Language Generation.
\newblock \emph{arXiv preprint arXiv:2506.11621}.

\bibitem[{Wang et~al.(2025{\natexlab{b}})Wang, Tang, Song, Wang, Guo, and Hong}]{wang2025linguistics}
Wang, X.; Tang, S.; Song, P.; Wang, S.; Guo, D.; and Hong, R. 2025{\natexlab{b}}.
\newblock Linguistics-vision monotonic consistent network for sign language production.
\newblock In \emph{ICASSP 2025-2025 IEEE International Conference on Acoustics, Speech and Signal Processing (ICASSP)}, 1--5. IEEE.

\bibitem[{Wang et~al.(2004)Wang, Bovik, Sheikh, and Simoncelli}]{wang2004image}
Wang, Z.; Bovik, A.~C.; Sheikh, H.~R.; and Simoncelli, E.~P. 2004.
\newblock Image quality assessment: from error visibility to structural similarity.
\newblock \emph{IEEE transactions on image processing}, 13(4): 600--612.

\bibitem[{Wei et~al.(2025)Wei, Sun, Ma, Hou, Juefei-Xu, He, Dai, Zhang, Li, Hou et~al.}]{wei2025mocha}
Wei, C.; Sun, B.; Ma, H.; Hou, J.; Juefei-Xu, F.; He, Z.; Dai, X.; Zhang, L.; Li, K.; Hou, T.; et~al. 2025.
\newblock Mocha: Towards movie-grade talking character synthesis.
\newblock \emph{arXiv preprint arXiv:2503.23307}.

\bibitem[{Wei, Yang, and Wang(2024)}]{wei2024aniportrait}
Wei, H.; Yang, Z.; and Wang, Z. 2024.
\newblock Aniportrait: Audio-driven synthesis of photorealistic portrait animation.
\newblock \emph{arXiv preprint arXiv:2403.17694}.

\bibitem[{Xu et~al.(2024{\natexlab{a}})Xu, Li, Su, Shang, Zhang, Liu, Wang, Yao, and Zhu}]{xu2024hallo}
Xu, M.; Li, H.; Su, Q.; Shang, H.; Zhang, L.; Liu, C.; Wang, J.; Yao, Y.; and Zhu, S. 2024{\natexlab{a}}.
\newblock Hallo: Hierarchical audio-driven visual synthesis for portrait image animation.
\newblock \emph{arXiv preprint arXiv:2406.08801}.

\bibitem[{Xu et~al.(2024{\natexlab{b}})Xu, Chen, Guo, Yang, Li, Zang, Zhang, Tong, and Guo}]{xu2024vasa}
Xu, S.; Chen, G.; Guo, Y.-X.; Yang, J.; Li, C.; Zang, Z.; Zhang, Y.; Tong, X.; and Guo, B. 2024{\natexlab{b}}.
\newblock Vasa-1: Lifelike audio-driven talking faces generated in real time.
\newblock \emph{Advances in Neural Information Processing Systems}, 37: 660--684.

\bibitem[{Zhang et~al.(2023{\natexlab{a}})Zhang, Wang, Zhang, Xu, Song, Xie, Luo, Tian, Guo, and Feng}]{zhang2023dream}
Zhang, C.; Wang, C.; Zhang, J.; Xu, H.; Song, G.; Xie, Y.; Luo, L.; Tian, Y.; Guo, X.; and Feng, J. 2023{\natexlab{a}}.
\newblock Dream-talk: Diffusion-based realistic emotional audio-driven method for single image talking face generation.
\newblock \emph{arXiv preprint arXiv:2312.13578}.

\bibitem[{Zhang et~al.(2018)Zhang, Isola, Efros, Shechtman, and Wang}]{zhang2018unreasonable}
Zhang, R.; Isola, P.; Efros, A.~A.; Shechtman, E.; and Wang, O. 2018.
\newblock The unreasonable effectiveness of deep features as a perceptual metric.
\newblock In \emph{Proceedings of the IEEE conference on computer vision and pattern recognition}, 586--595.

\bibitem[{Zhang et~al.(2023{\natexlab{b}})Zhang, Cun, Wang, Zhang, Shen, Guo, Shan, and Wang}]{zhang2023sadtalker}
Zhang, W.; Cun, X.; Wang, X.; Zhang, Y.; Shen, X.; Guo, Y.; Shan, Y.; and Wang, F. 2023{\natexlab{b}}.
\newblock Sadtalker: Learning realistic 3d motion coefficients for stylized audio-driven single image talking face animation.
\newblock In \emph{Proceedings of the IEEE/CVF conference on computer vision and pattern recognition}, 8652--8661.

\bibitem[{Zhou et~al.(2020)Zhou, Han, Shechtman, Echevarria, Kalogerakis, and Li}]{zhou2020makelttalk}
Zhou, Y.; Han, X.; Shechtman, E.; Echevarria, J.; Kalogerakis, E.; and Li, D. 2020.
\newblock Makelttalk: speaker-aware talking-head animation.
\newblock \emph{ACM Transactions On Graphics}, 39(6): 1--15.

\end{thebibliography}

\end{document}